# Analysis of the quotation corpus of the Russian Wiktionary

Alexander Smirnov[1], Tatiana Levashova[1], Alexey Karpov[2], Irina Kipyatkova[1], Andrey Ronzhin[2], Andrew Krizhanovsky[3] and Nataly Lugovaya[3]

[1]Institution of the Russian Academy of Sciences St.Petersburg Institute for Informatics and Automation RAS

[2]Saint-Petersburg State University, Department of Phonetics, Russia

[3]Institute of Applied Mathematical Research of the Karelian Research Centre of the Russian Academy of Sciences

## ABSTRACT

The quantitative evaluation of quotations in the Russian Wiktionary was performed using the developed Wiktionary parser. It was found that the number of quotations in the dictionary is growing fast (51.5 thousands in 2011, 62 thousands in 2012). These quotations were extracted and saved in the relational database of a machine-readable dictionary. For this database, tables related to the quotations were designed. A histogram of distribution of quotations of literary works written in different years was built. It was made an attempt to explain the characteristics of the histogram by associating it with the years of the most popular and cited (in the Russian Wiktionary) writers of the nineteenth century. It was found that more than one-third of all the quotations (the example sentences) contained in the Russian Wiktionary are taken by the editors of a Wiktionary entry from the Russian National Corpus.

## 1. INTRODUCTION

The progress of computer technologies provides a basis for a new type of dictionaries. This is an online dictionary, where any interested person can take part in the development. In comparison with the traditional lexicography, on the one hand, this way of organizing collective work provides obvious advantages (high intensities of work, the possibility to discuss and correct the articles in real time at any stage of work). On the other hand, there is a high possibility that gaps can be presented in the source material and gaps can be found in the dictionary itself.

One of the possible solutions to this problem is to develop a special software tool that can analyse an online dictionary at any stage of the development. Some possible solutions to this problem will be presented in the paper on the basis of the analysis of the quotations which illustrate the words meaning in the Russian Wiktionary.

Therefore, the goals of this work are (1) to construct the quotation corpus from the online dictionary, (2) to analyse the chronological distribution of the quotation corpus in the time period of 1750 to 2012. This time period includes years with more than 10 quotations which refer to this year in the dictionary.

The definition given in the paper [5]Narinyany2001 for the dictionaries of the future can be applied to the Wiktionary – it is "an intellectual computer dictionary which combines thesaurus, lexicon and phraseological dictionary, and it is integrated with dictionaries in other languages". Indeed, the Wiktionary is a multilingual and multifunctional dictionary and thesaurus.

The Wiktionary combines a glossary and a defining, grammatical, etymological, and translation dictionaries. Consequently, the Wiktionary contains not only word's definitions, semantically related words (synonyms, hypernyms, etc.), translations, but also the pronunciations (phonetic transcriptions, audio files), hyphenations, etymologies, quotations, parallel texts (quotations with translations), figures (which illustrate meaning of the words).

The advantages of the Wiktionary are the huge volume of data and the great variety of the lexicographical material. It was shown in the papers [3]Krizhanovsky2012, [4]Meyer2012 that the size of the German Wiktionary is comparable with thesauri *GermaNet* and *OpenThesaurus*, and that the size of the English Wiktionary exceeds the size of the *WordNet*. It is possible that the freely available[1] dictionary in Russian comparable with the Russian Wiktionary by size and the broad range of lexicographical material does not exist at this moment.

A word's definition in the Wiktionary is accompanied by quotations, which illustrate the meaning by surrounding context. Quotations can include reference to the source (book, newspaper, blog) with a date of publication or writing.

The analysis of the dates of literary works, which are used as a source of quotations is the goal of the paper. The experiments were carried out on the corpus of quotations build on the basis of the Russian Wiktionary. The database of quotation corpus is a part of the machine-readable Wiktionary, which is an open-source project.[2]

## 2. THE FRAMEWORK OF THE MACHINE-READABLE WIKTIONARY

The conception of the machine-readable Wiktionary is flexible in relation to input data, but it is strict and formal to output data.

*Input data.* This conception suggests that different wiktionaries can have different structure of the article (e.g. different names of the article sections and different order of sections), which should be taken into account by a parser of wiktionaries. Moreover, even within one Wiktionary the structure of the article can change with time as new sections appear and templates vary and modify. Therefore there is need for a flexible and modular framework in order to parse so much "live" and various wiktionaries (Figure 1). The specific properties of different wiktionaries are taken into account in the submodules "*ruwikt*" and "*enwikt*" in the module "*Data extraction*" in Figure 1.

*Output data*. The data extracted from a Wiktionary are stored in the database of the machine-readable dictionary. The result databases filled by the parser have

1 Freely available for readers and editors.

2 The machine-readable Wiktionary is available at http://code.google.com/p/wikokit/

identical structure independent on the source wiktionary. This provides compatibility of different machine-readable wiktionaries with external applications.

The adding of new wiktionaries in a system can be done in practice, because many parts of the parser are already developed and do not depend on a specific wiktionary.

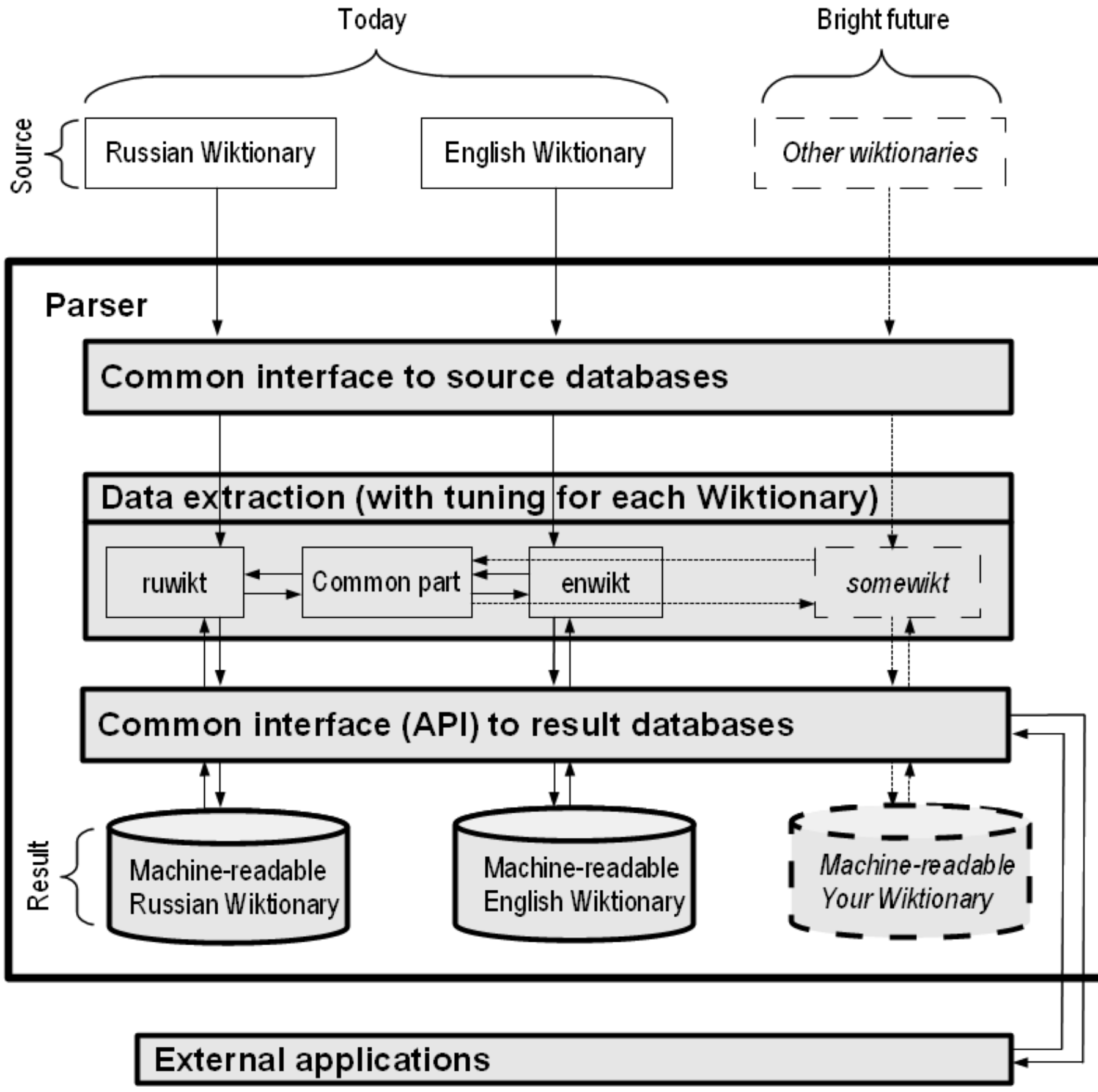

**Figure 1. Machine-readable Wiktionary framework**

There are the following common parts:

- Common application programming interface (API) to the source databases (read data).
- “Common part” of the module “Data extraction” (Figure 1). For example it contains (1) the language codes in accordance with an international standard for language codes ISO 639, (2) names of languages in English and in Russian; now the parser recognizes 370 languages and codes in the English Wiktionary and 274 in the Russian Wiktionary.

- Common API to the result databases of the machine-readable wiktionaries (read / write data).

## 3. THE ARCHITECTURE OF THE DATABASE OF QUOTATION CORPUS

The database of quotation corpus is a part of the relational database of the machine-readable Wiktionary presented in the paper [2]Krizhanovsky2010.

The following fields of the quotation template are recognized and added to the database during the extraction of semistructured data of the Wiktionary by the parser (Figure 2):

- The text of the quotation (stored in the field *text* of the table *quote*).
- The translation into Russian (the table *quot_translation*).
- The transcription of the quotation (the table *quot_transcription* is reserved for the English Wiktionary, it is not used in the Russian Wiktionary).

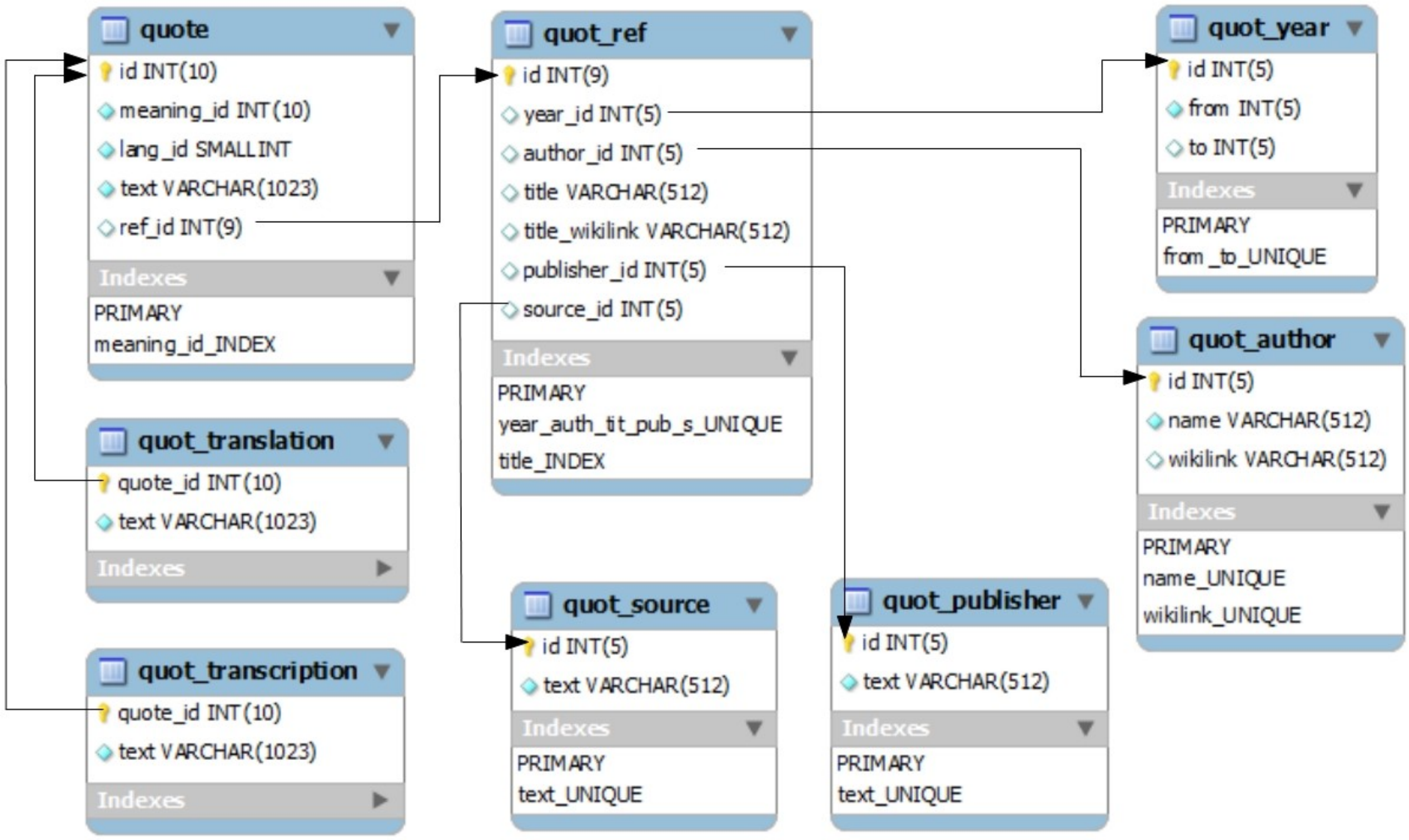

**Figure 2. Tables and relations related to quotations in the database of the machine-readable Wiktionary**

- The quotation's reference joins the following elements in the table *quot_ref*:
  - A title of the source (the field *title* of the table *quot_ref*).
  - An author of the source (the table *quot_author*).
  - A publisher (the table *quot_publisher*).
  - Publication date (the table *quot_year*).
  - The name of a resource or a corpus, where the quotation was taken from (the table *quot_source*).

# 4. DATABASE QUERIES

It is possible to construct various SQL-queries using tables related to quotes (Figure 2). For example, using a few queries, one can solve the following search task: *get a list of quotations in English, which refer to books written during more than one year.*[3]

[3] See these queries: http://code.google.com/p/wikokit/wiki/MRDQuote#SQL_queries

This task should be solved step by step:

1) Get a list of quotations in English. As of March 2012, there are 1355 quotes in English in the Russian Wiktionary.
2) Get a sublist of quotations with non-empty reference (a source). There are 222 quotations where “*ref_id”* is not NULL in the table “*quote”*.
3) Get a subsublist of quotations, which contain a date in the reference. There are 123 quotations with years.
4) At last, get a list of quotations, which contain a range of years in the reference. I.e. the value of the field “*to”* is greater than “*from*” in the table *quot_year* (Figure 2). As a result the seven quotes were found, see Table 1.

**Table 1. English quotations from the Russian Wiktionary, which refer to books written during more than one year**

| N | Entry | author | title | from | to |
|---|---|---|---|---|---|
| 1 | Moscow[4] | Андрей Платонов | Эфирный тракт | 1926 | 1927 |
| | **Moscow** awakened and screamed with trams. … The summer sun rejoiced over the full-blooded land, and two men appeared before the gaze of a new **Moscow** — a wonderful city of powerful culture, stubborn labor and intelligent happiness. | | | | |
| | **Москва** проснулась и завизжала трамваями. … Летнее солнце ликовало над полнокровной землёй, и взорам двух людей предстала новая **Москва** — чудесный город могущественной культуры, упрямого труда и умного счастья. | | | | |
| 2 | cacophony | H. P. Lovecraft | Herbert West: Reanimator | 1921 | 1922 |
| | Not more unutterable could have been the chaos of hellish sound if the pit itself had opened to release the agony of the damned, for in one inconceivable **cacophony** was centered all the supernal terror and unnatural despair of animate nature. | | | | |
| 3 | hoarder | – | [Central News autocue data.] 3623 s-units. | 1985 | 1994 |
| | The picture was owned by antiques **hoarder** Ronnie Summerfield who died three years ago leaving a collection valued at millions of pounds. | | | | |
| 4 | hoarder | – | The Economist. 3341 s-units. | 1985 | 1994 |
| | The biggest official gold **hoarder** by far is America, which holds 27,9 % of the world's central-bank gold reserves. | | | | |
| 5 | order | Charles Dickens | Oliver Twist | 1837 | 1839 |
| | In pursuance of this determination, little Oliver, to his excessive astonishment, was released from bondage, and **ordered** to put himself into a clean shirt. | | | | |
| | Во исполнение этого решения маленького Оливера, крайне удивленного, выпустили из заточения и **приказали** надеть чистую рубашку. | | | | |
| 6 | order | Charles Dickens | Oliver Twist | 1837 | 1839 |
| | Oliver was **ordered** into instant confinement; and a bill was next morning pasted on the outside of the gate, offering a reward of five pounds to anybody who would take Oliver Twist off the hands of the parish. | | | | |
| | Было **предписано** немедленно отправить Оливера в заточение; а на следующее утро к воротам было приклеено объявление, что любому, кто пожелает освободить приход от Оливера Твиста, предлагается вознаграждение в пять фунтов. | | | | |
| 7 | practitioner | Чарльз Диккенс | The Posthumous Papers of the Pickwick Club | 1836 | 1837 |
| | These sequestered nooks are the public offices of the legal profession, where writs are issued, judgments signed, declarations filed, and numerous other ingenious machines put in motion for the torture and torment of His Majesty's liege subjects, and the comfort and emolument of the **practitioners** of the law. | | | | |
| | В этих изолированных уголках помещаются официальные конторы адвокатов, где выдают судебные приказы, заверяют судебные решения, подшивают жалобы истцов и приводят в действие многие другие хитроумные приспособления, изобретенные для мук и терзаний подданных его величества и для утешения и обогащения **служителей** закона. | | | | |

The column "*entry*" in Table 1 contains the headword of the Wiktionary article. The quotation is placed in the row below and the word in question is marked by **bold** font in the quote. If there is the translation of this quote into the Russian then it is presented in the row below. The author name, the title of the source book and the publication (or writing) date (in years) are given in the columns "*author*", "*title*", "*from*", "*to*".

[4] See entry "Moscow" in the Russian Wiktionary: http://ru.wiktionary.org/wiki/Moscow

# 5. EXPERIMENTS

## 5.1 Corpus of quotations

The corpus of quotations was built on the basis of the Russian Wiktionary, dump as of March 25, 2012. It was constructed with the help of the developed Wiktionary parser [2]Krizhanovsky2010. It includes 62 thousand quotations (51.5 thousand in 2011). It is significant that 52 thousand quotations (84% of the whole number of quotations) illustrate the Russian words (82% in 2011).

In the Russian Wiktionary 23.8 thousand quotations (38.35% of the whole number) have a reference to the source (17 thousand quotations with references in 2011, i.e. 33%). The main source of quotations in the Russian Wiktionary is the *Russian National Corpus* [1]Apresjan2006. There are 94.15% quotations (of the whole number of quotations with references) which refer to the Russian National Corpus.

## 5.2 Publication date: analysis and hypothesis

In this study publication dates indicated in the sources of quotations are under investigation. The table *quot_year* contains two fields "*from*" and "*to*" having only integer values (Figure 2), which indicate the years of publication of the source. If the work was published or written during one year, then both fields have equal values. The number of unique pairs (start year "*from*", finish year "*to*") is 862 in the Russian Wiktionary (it equals to the number of records in the table *quot_year*).

In order to calculate a number of quotations for each year the algorithm similar to the known game "Tetris" was used (Figure 3). The algorithm traverses all quotations, if the quotation contains a year or a range of years then the number of quotations for years in this range are incremented.

For example, there are years 1927 and 1945-1955 in the following quotations of entries «Возрождение» and «танцевать» in the Russian Wiktionary:

- Она бежала мимо парчовых кресел итальянского **Возрождения**, мимо голландских шкафов, мимо большой готической кровати с балдахином на чёрных витых колоннах. *Илья Ильф, Евгений Петров, «Двенадцать стульев», **1927** г.*
- Она никогда не могла предположить, что он так хорошо **танцует**. *Б. Л. Пастернак, «Доктор Живаго», **1945—1955** г.*

These quotations are presented in Figure 3 in the form of bricks on the abscissa in 1927 and in the range 1945-1955. Suppose that during the traversal, a quotations from the source written in 1950-1957 is reached. Then the value of the histogram at 1950-1955 in Figure 3 is two (quotations) and at 1956-1957 is one (quotation).

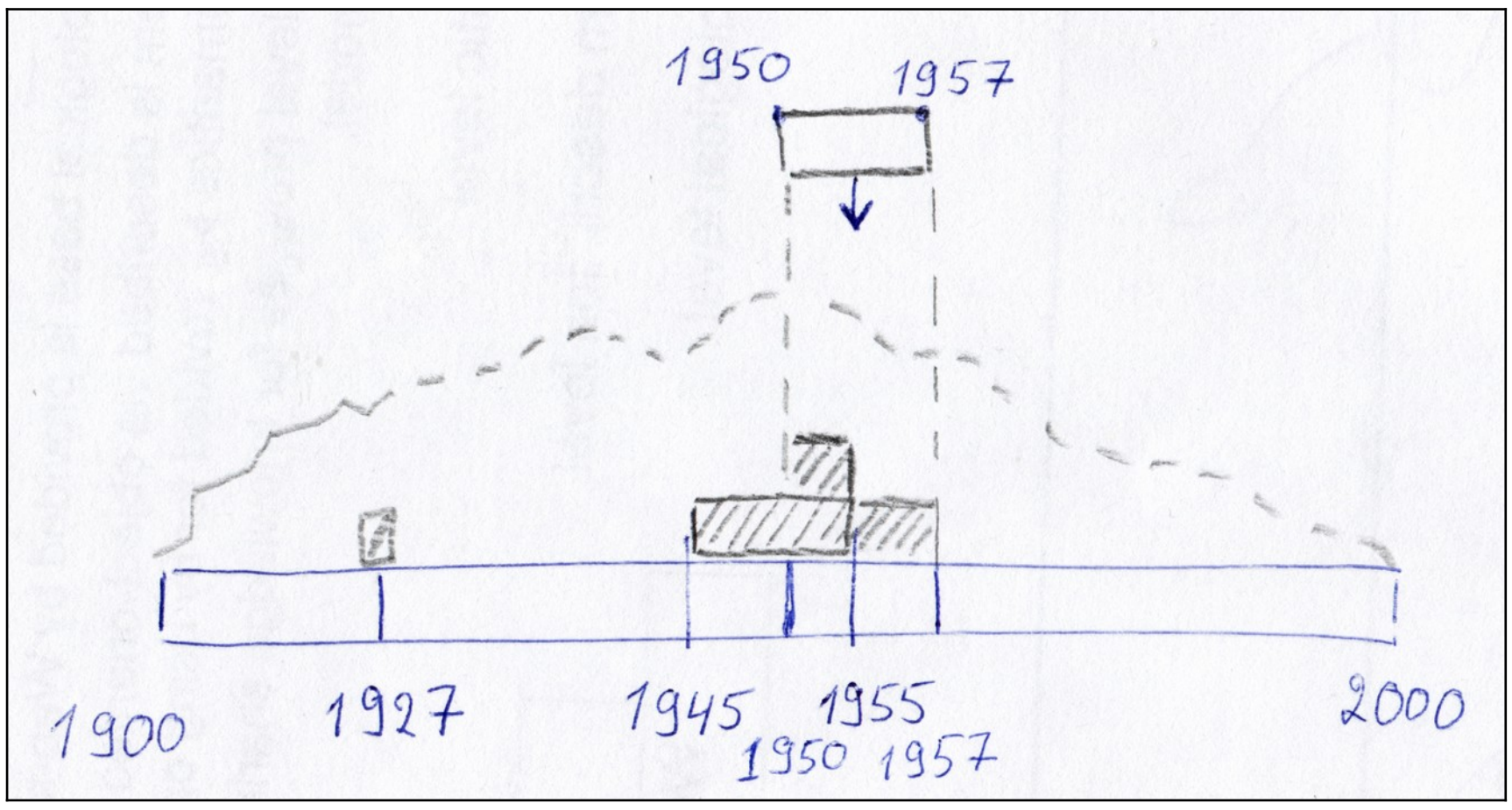

**Figure 3. An idea of a calculation of the number of quotes for each year in the online dictionary (histogram)**

The traversal of 26596 quotations (which contains date) makes possible to build the following histogram (Figure 4), which relates the number of quotations and the source's publication date in the range 1750-2012.[5]

[5] The source data for Figure 4 is available at http://ru.wiktionary.org/Участник:AKA%20MBG/Статистика:Цитаты%20(дата)

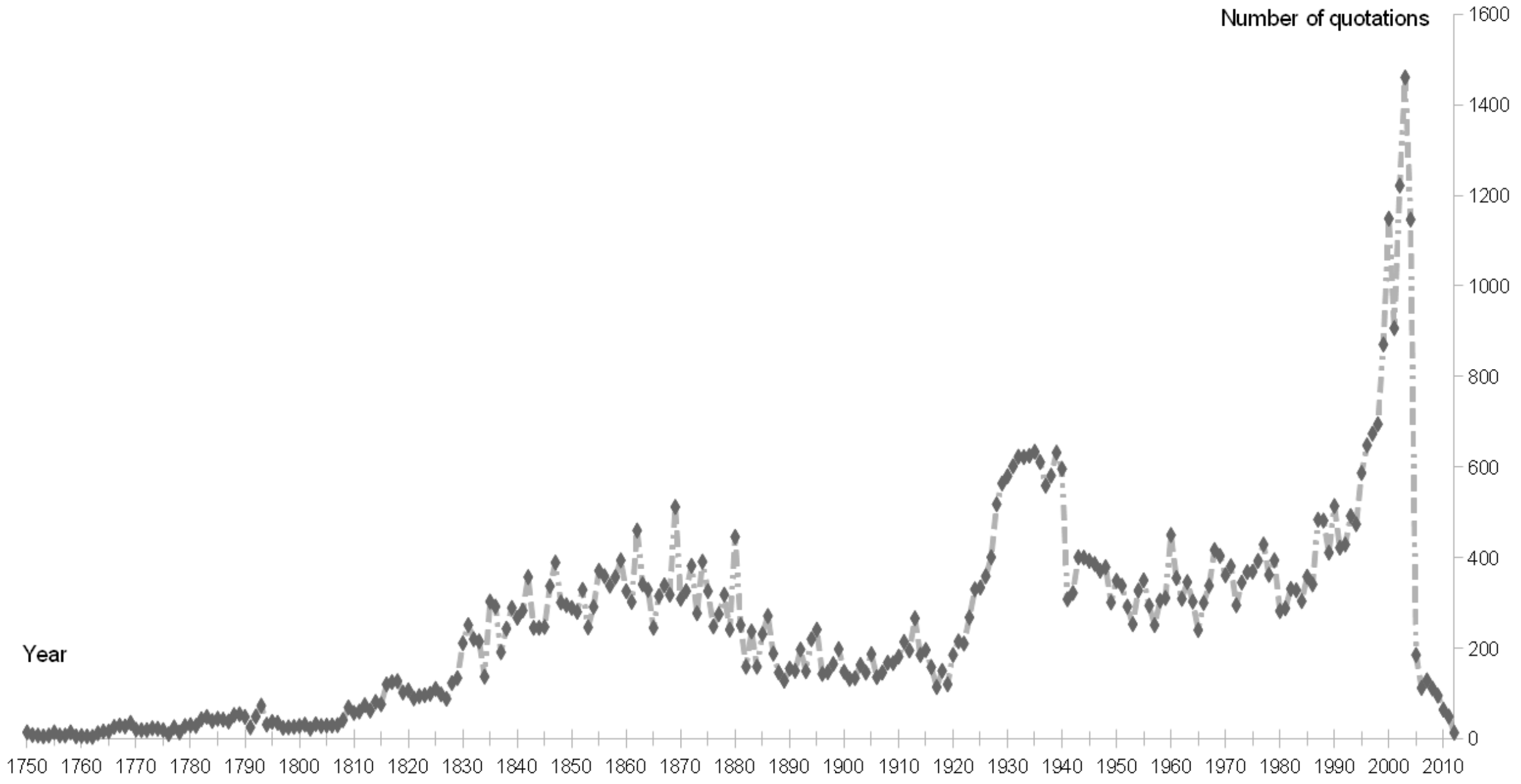

**Figure 4. The dependence of the number of quotations with respect the source's publication date**

The peak number of quotations in the 2000s might be explained by a relatively high number of newspapers and journals available at the Russian National Corpus within this time range. Since RNC is the main source of quotations for the Russian Wiktionary (see above).

In order to understand the relatively high number of quotations in Figure 5 in the time range from the 1830s to the 1880s, the contribution of the most cited in the Russian Wiktionary writers will be analysed.

The writers with the highest number of quotations in the Russian Wiktionary are listed in the column "Author" in Table 2. The second and third columns demonstrate the fast growing size of the dictionary in a number of quotations for these writers in 2011 and 2012.

The main source of quotations in the Russian Wiktionary is the Russian National Corpus hence there is a column labeled "Publication in Russian National Corpus", which provides the years of the first and last publications of the author presented in the corpus. For the same periods the total numbers of quotations were counted (column "Total quotes…"). The last column is the ratio of the number of quotations of the author (third column) to the total numbers of quotations for the periods, when the publications of the author is presented in the corpus (the last but one column).

**Table 2. The most popular authors in the Russian Wiktionary**

| Author | Number of quotes (2011) | Number of quotes (2012) | Publication in Russian National Corpus | Total quotes in Wiktionary (within this time range) | Contri bution % (2012) |
|---|---|---|---|---|---|
| Anton Chekhov | 716 | 931 | **1880**-1904 | 4704 | 19,8% |
| Leo Tolstoy | 529 | 710 | 1852-**1910** | 14954 | 4,8% |
| Alexander Pushkin | 520 | 627 | 1815-1836 | 3217 | 19,5% |
| Fyodor Dostoyevsky | 500 | 776 | 1846-1881 | 11853 | 6,6% |
| Ivan Turgenev | 457 | 697 | 1846-1882 | 12012 | 5,8% |
| Nikolai Gogol | 321 | 473 | 1831-1847 | 4511 | 10,5% |
| Nikolai Leskov | 245 | 386 | 1862-1894 | 9039 | 4,3% |
| Mikhail Bulgakov | 207 | 267 | 1920-1940 | 10049 | 2,7% |
| Arkady and Boris Strugatskye | 171 | 225 | 1964-1979 | 5699 | 4,0% |
| Viktor Astafyev | 142 | 199 | 1967-2001 | 16327 | 1,2% |

The total number of quotations of the first seven authors in the period 1815-1910 (*Chekhov – Leskov* in Table 2) is 22429, it is 20.5%, i.e. one fifth part of the whole amount of quotations in this period in the Russian Wiktionary.

It is possible that the high citations of these writers is the reason of the peak in Figure 5 in the time range from the 1830s to the 1880s,

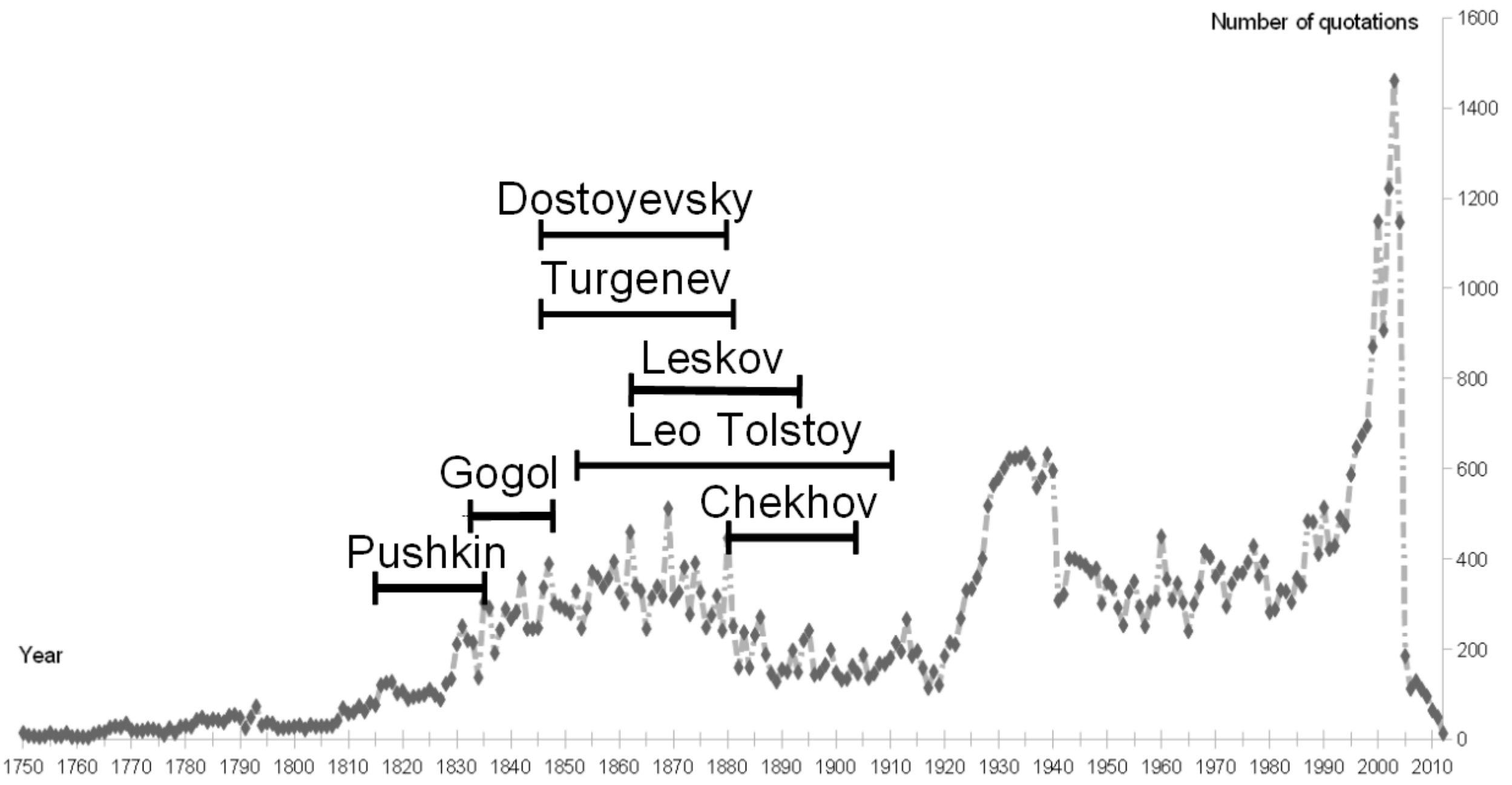

**Figure 5. The dependence of the number of quotations with respect the source's publication date and the years of literary activity of the most cited in the Russian Wiktionary writers**

The open question remains, what authors contribute to the peak in Figure 5 in the period 1920s – 1940s before the war?

## 5.3 Distribution for centuries

The analysis revealed that the earliest quotations in the Russia Wiktionary are dated:

- 70 BC, Cicero, "Against Verres", Latin, the entry "asylum" [6].
- 1076, «Изборник Святослава» (Svyatoslav's Miscellanies), Old East Slavic, the entry «воинъ»[7].
- 1364, Guillaume de Machaut, "Dit de la Marguerite", Old French, the entry "chançon"[8].

In the course of experiments the distribution of quotes from the Russian Wiktionary dating from 17th to 21st century was made (Figure 6). The 21st century corresponds to the range 2000-2012, inclusively.

---

6 See http://ru.wiktionary.org/wiki/asylum

7 See http://ru.wiktionary.org/wiki/%D0%B2%D0%BE%D0%B8%D0%BD%D1%8A

8 See http://ru.wiktionary.org/wiki/chan%C3%A7on

It could be seen that each subsequent century contains more quotations than the previous one. Probably, this tendency will remain, since the first 12 years of this century already have given 10% of the whole number of quotations in the dictionary.

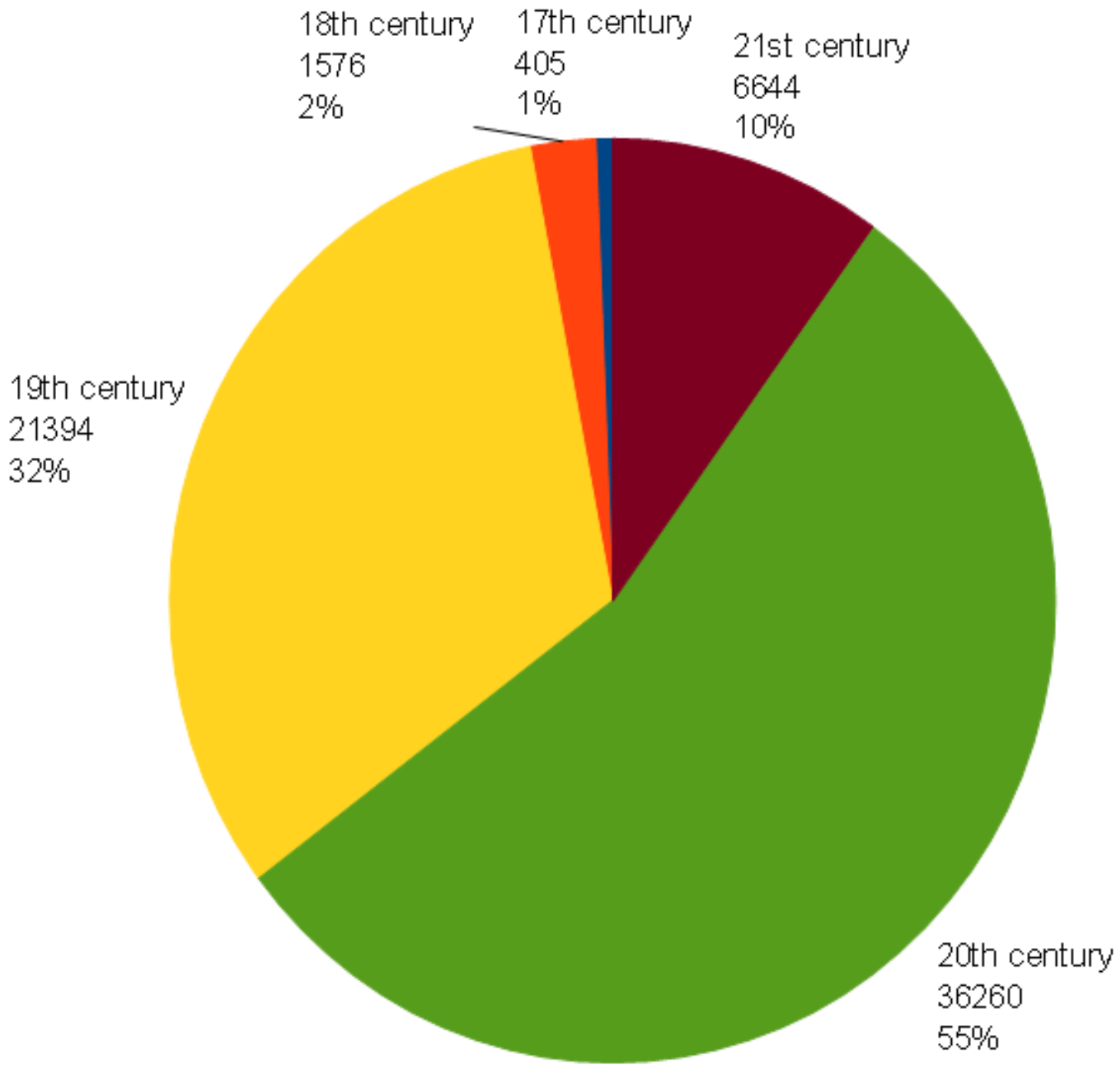

**Figure 6. The distribution of Wiktionary quotes dating from 17th to 21st century (21st century corresponds to the range 2000-2012)**

# 6. CONCLUSION

In this paper the framework of the machine-readable Wiktionary was designed, which emphasizes the possibility to add new wiktionaries to the system.

The architecture of the database of quotation corpus was described. An exemplary search task (to get a list of quotations in English, which refer to books written during more than one year) was solved.

The characteristics of corpus of quotations constructed on the basis of the Russian Wiktionary were investigated. It was found that the number of quotations in the dictionary grows fast (51.5 thousands in 2011, 62 thousands in 2012).

The histogram which relates the number of quotations and the source's publication date in the range 1750-2012 was created. It was made an attempt to explain the characteristics of the histogram by associating it with the years of the most popular and cited (in the Russian Wiktionary) writers of the nineteenth century: Anton Chekhov, Leo Tolstoy, Alexander Pushkin, Fyodor Dostoyevsky, Ivan Turgenev, Nikolai Gogol, and Nikolai Leskov.

## 7. ACKNOWLEDGMENTS

Some parts of the research were carried out under projects funded by grants # 11-01-00251, # 12-01-00481 and # 12-07-00070 of the Russian Foundation for Basic Research, grant # 12-04-12062 of the Russian Foundation for Humanities and project of the research program “Intelligent information technologies, mathematical modelling, system analysis and automation” of the Russian Academy of Sciences. Some parts of this work were supported by the Ministry of Education and Science of Russian Federation (The Russian Federal Targeted Program "R&D in Priority Fields of S&T Complex of Russia for 2007-2013", Contract No. 07.514.11.4139).